\documentclass[letterpaper, 10 pt, conference]{ieeeconf}  % Comment this line out if you need a4paper

\IEEEoverridecommandlockouts                              % This command is only needed if 
\usepackage{graphicx}
\usepackage{paralist}
\usepackage{epsfig} % for postscript graphics files
\usepackage{mathptmx} % assumes a new font selection scheme installed
\usepackage{times} % assumes a new font selection scheme installed
\usepackage{amsmath} % assumes amsmath package installed
\usepackage{amssymb}  % assumes amsmath package installed

\usepackage{subcaption}
\usepackage{multirow}
\usepackage{hyperref}
\usepackage{xcolor}
\usepackage{cite}
\usepackage{colortbl}
\usepackage{dirtytalk}
\usepackage[normalem]{ulem}
\usepackage{svg}

\title{\LARGE \bf
Comprehensive Robotic Cholecystectomy Dataset (CRCD)$^{\dagger}$: Integrating Kinematics, Pedal Signals, and Endoscopic Videos
}

\author{Ki-Hwan Oh$^{1*}$, Leonardo Borgioli$^{1*}$, Alberto Mangano$^{2}$, Valentina Valle$^{2}$, Marco Di Pangrazio$^{3}$,\\
Francesco Toti$^{4}$, Gioia Pozza$^{5}$, Luciano Ambrosini$^{2}$, Alvaro Ducas$^{2}$, Milo\v s \v Zefran$^{1}$, Liaohai Chen$^{2}$\\
and Pier Cristoforo Giulianotti$^{2}$% <-this % stops a space
\thanks{$^{\dagger}$ \url{https://github.com/Borgioli/crcd_ros}}% <-this % stops a space
\thanks{$^{*}$ First two authors contributed equally to this work.}
\thanks{$^{1}$ Robotics Lab,  Department of Electrical and Computer Engineering, College of Engineering, University of Illinois Chicago, Chicago, IL 60607, USA.}%
\thanks{$^{2}$ Surgical Innovation and Training Lab,  Department of Surgery, College of Medicine, University of Illinois Chicago, Chicago, IL 60607, USA.}%
\thanks{$^{3}$ School of Medicine and Surgery, University of Modena and Reggio Emilia, Modena, Italy.}%
\thanks{$^{4}$ Department of Medicine and Surgery, University of Milan, Milano, Italy.}%
\thanks{$^{5}$ Department of Surgery, Lugano Regional Hospital, Ente Ospedaliero Cantonale (EOC), Lugano, Switzerland.}%
}

\begin{document}

\maketitle
% \thispagestyle{empty}
% \pagestyle{empty}

%%%%%%%%%%%%%%%%%%%%%%%%%%%%%%%%%%%%%%%%%%%%%%%%%%%%%%%%%%%%%%%%%%%%%%%%%%%%%%%%
\begin{abstract}

%Robotic surgery promises enhanced precision and adaptability over traditional surgical methods. It also offers the possibility of automating surgical interventions, resulting in reduced stress on the surgeon, improved surgical outcomes, and lower costs. One emerging approach in robotics is reinforcement learning, which is gaining prominence in the field. However, this approach has not been explored in the domain of surgical robotics, primarily due to the lack of kinematic data. This research aims to present the first dataset, to our knowledge, by merging time-stamped videos with kinematic data collected from a da Vinci Research Kit (dVRK) during cholecystectomy procedures performed by five surgeons on ex-vivo pig livers.

In recent years, the potential applications of machine learning to Minimally Invasive Surgery (MIS) have spurred interest in data sets that can be used to develop data-driven tools. This paper introduces a novel dataset recorded during ex vivo pseudo-cholecystectomy procedures on pig livers, utilizing the da Vinci Research Kit (dVRK). Unlike current datasets, ours bridges a critical gap by offering not only full kinematic data but also capturing all pedal inputs used during the procedure and providing a time-stamped record of the endoscope's movements.
Contributed by seven surgeons, this data set introduces a new dimension to surgical robotics research, allowing the creation of advanced models for automating console functionalities. Our work addresses the existing limitation of incomplete recordings and imprecise kinematic data, common in other datasets. By introducing two models, dedicated to predicting clutch usage and camera activation, we highlight the dataset's potential for advancing automation in surgical robotics. The comparison of methodologies and time windows provides insights into the models' boundaries and limitations.

\end{abstract}

%%%%%%%%%%%%%%%%%%%%%%%%%%%%%%%%%%%%%%%%%%%%%%%%%%%%%%%%%%%%%%%%%%%%%%%%%%%%%%%%

\section{Introduction}

The training of state-of-the-art models requires the development of extensive datasets. In recent years, considerable efforts have been made to establish sizable public datasets for surgical procedures, featuring comprehensive annotations from experts. The creation of extensive data sets specifically focused on the execution of surgical tasks using robotic systems is an important step toward furthering these advances. The datasets offer a comprehensive portrayal of the surgeon's actions, encompassing both kinematic and dynamic data, alongside recorded videos.

Most of the data sets focus on segmentation of the instruments~\cite{bouget2015, ross2020robust} and/or organs~\cite{allan20202018, hong2020cholecseg8k, carstens2023dresden} captured by the endoscope during surgical procedures. For example, \cite{twinanda2016endonet} is a video data set with instrument segmentations that also includes the labeling of different phases of the cholecystectomy (surgical removal of the gallbladder) procedure (similar to Section~\ref{section:task_desc}), and was used to train EndoNet~\cite{twinanda2016endonet} which predicts the presence of instruments and recognizes the current surgical phase. However, these datasets do not include kinematic data. This makes it difficult to estimate the 3D position of the detected instrument and calculate its distance to the tissues. Moreover, kinematic data has been reported to help improve tool segmentation~\cite{su2018, da2019self}.

Few state-of-the-art datasets include kinematic data, as illustrated in~\cite{dataset_review}. For example, in~\cite{kinematic_dataset}, they recorded the kinematics of the controllers and the surgical robot arms while recording the video from non-endoscopic stereo cameras. However, the cameras were in a fixed location, and the way the images were captured was different from what was shown in endoscopic videos. In addition, they performed basic training tasks such as moving a peg or following a wire on a board, as opposed to real surgical procedures. More advanced tasks such as suturing and knottying were performed on the JIGSAWS~\cite{gao2014jhu} dataset, but it was a toy experiment and not applied to real tissues. In~\cite{colleoni2020synthetic}, kinematics was recorded, but it was only used to improve the instrument segmentation data set to be more robust with respect to different background tissues, and the movements were not related to surgical procedures.

In addition, one of the significant but trivial interaction signals is ignored in the existing datasets, which are the pedals of the robot surgery system. Surgeons frequently use the pedals to stop the robot arms and move the endoscope, and to apply mono/bipolar power to the instrument to dissect tissues. Analyzing these interactions and automating these secondary tasks is the key to alleviating the stress and burden on surgeons during prolonged surgical interventions. 

To address the shortcomings observed in the previously released datasets, we decided to record cholecystectomy procedures using a coupled set of videos, kinematics data, and pedal signals. 
Cholecystectomy was chosen since it is one of the popular and standard laparoscopic procedures~\cite{cullen2009ambulatory, harrell2005minimally}. The same applies to robotic cholecystectomy which has been gradually (mainly due to perceived higher costs) increasing in popularity~\cite{strosberg2017retrospective}. The robotic cholecystectomy procedures are similar to the laparoscopic approach, and the details of the procedure are described in Section~\ref{section:task}. 

\section{Dataset Composition}

\subsection{Stereo Endoscopic Images}

\begin{figure}[t]
    \centering
    \begin{subfigure}[b]{0.49\columnwidth}
        \centering
        \includegraphics[width=\linewidth]{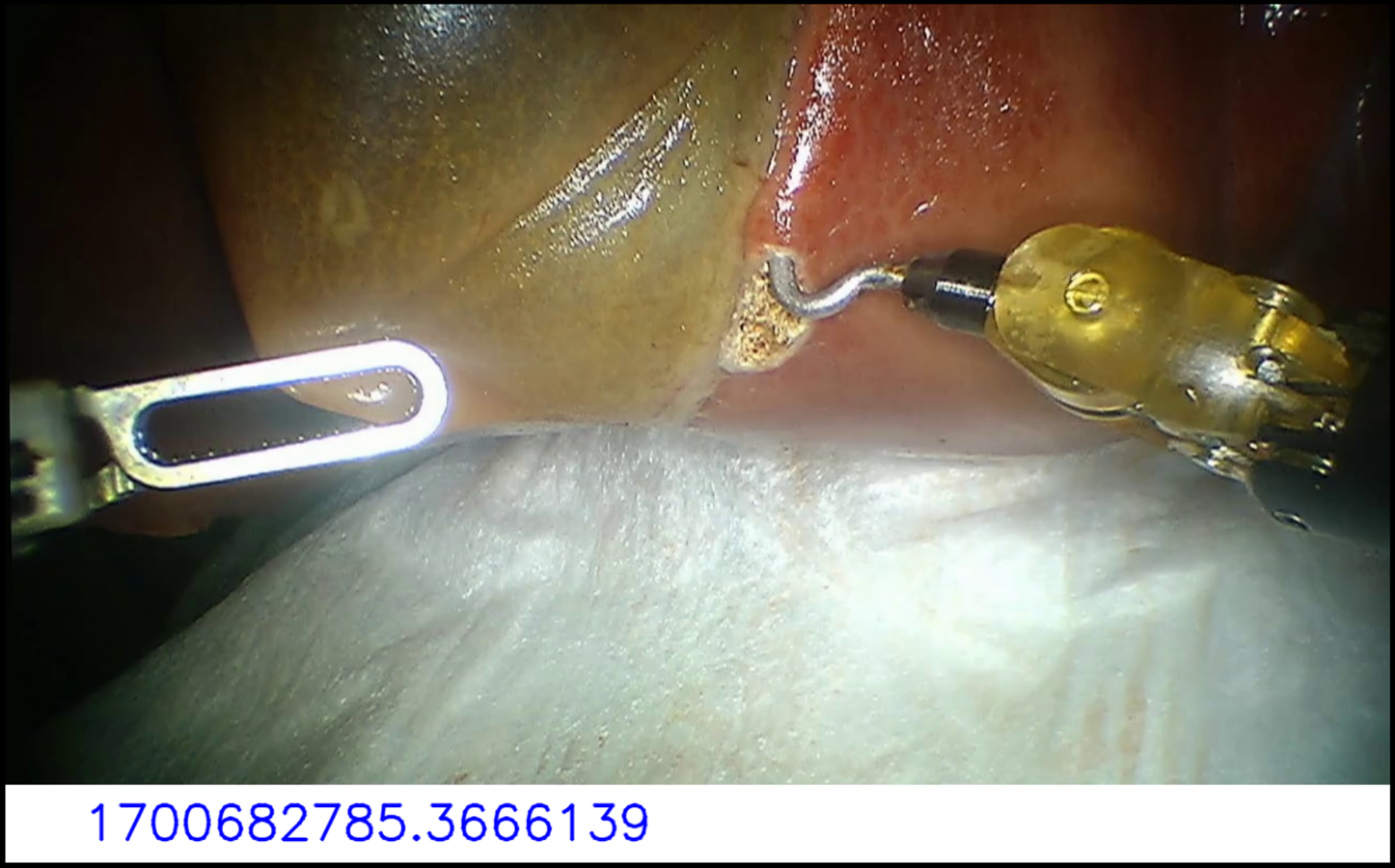}
        \caption{Left endoscope.}
        % \label{fig:left}
    \end{subfigure}
    \begin{subfigure}[b]{0.49\columnwidth}
        \centering
        \includegraphics[width=\linewidth]{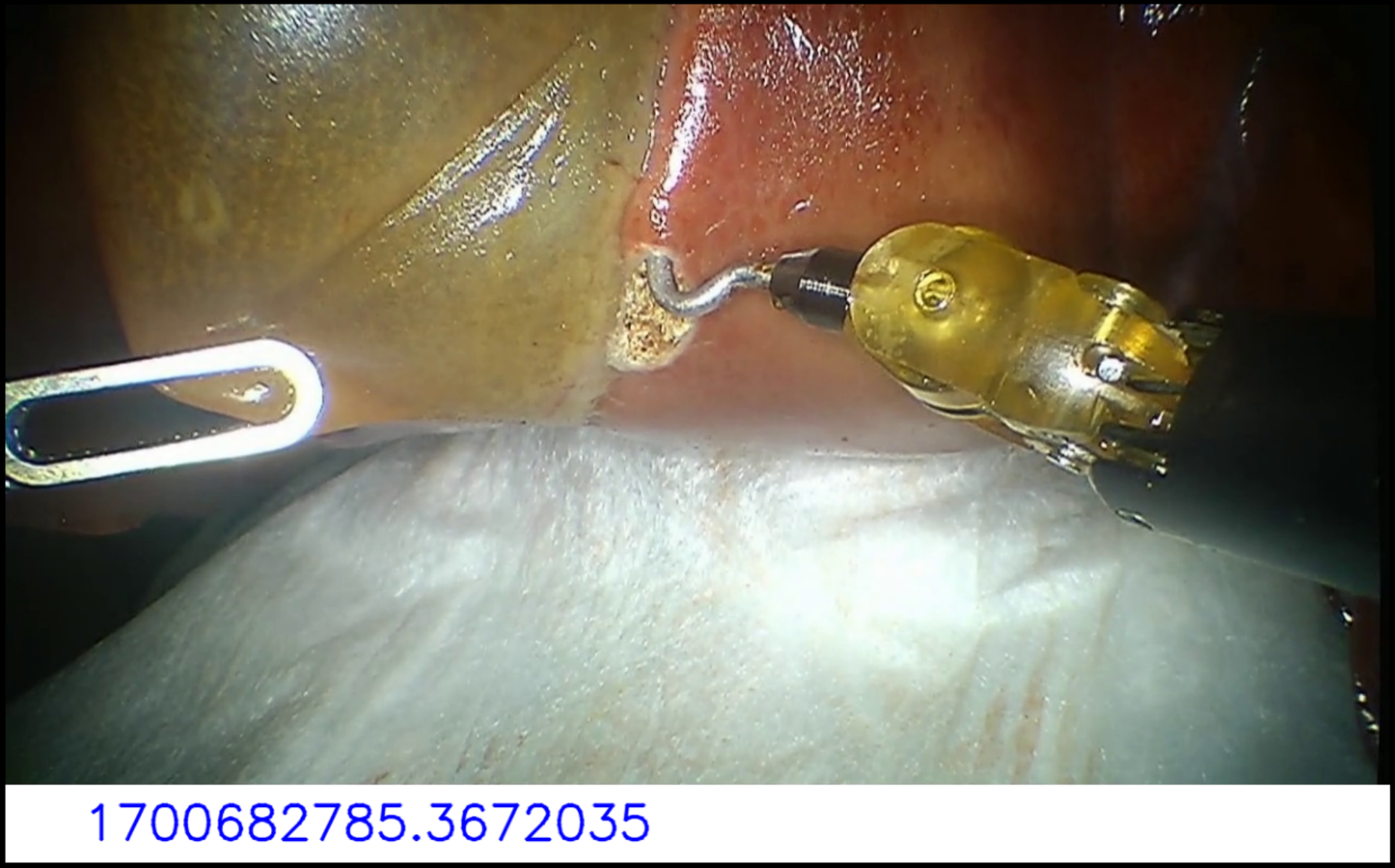}
        \caption{Right endoscope.}
        % \label{fig:right}
    \end{subfigure}
    \caption{Sample of the stereo endoscopic images.}
    \label{fig:stereo}
\vspace{-5mm}
\end{figure}

This study utilizes the first-generation da Vinci surgical system integrated with the dVRK. In contrast to traditional configurations, we selected the Si model endoscope due to its superior image quality and significantly reduced noise characteristics. The stereo endoscope cameras are calibrated as described in OpenCV~\cite{opencv_library}, based on the approaches presented in~\cite{zhang2000} and~\cite{caltoolbox}, finding the intrinsic and extrinsic parameters for each camera. The dataset includes distortion parameters, intrinsic camera matrix, rectification matrix, and projection matrix for the left and right endoscopes. These parameters facilitate the recovery of 3D point clouds from the recorded videos.

Individual images from each camera are recorded separately, featuring an additional timestamp placed at the bottom of the image, as illustrated in Fig.~\ref{fig:stereo}. The timestamps are from the Robot Operating System (ROS)~\cite{ros} and can be extracted utilizing optical character recognition (OCR) engines such as the Tesseract~\cite{tesseract}. With these timestamps, one could find the corresponding kinematics and pedal signals from the dataset. Each video was recorded with a rate of 60 frames per second and a resolution of $1280 \times 720$ pixels. The videos are encoded in AVC1 four-character code (FourCC) and compressed to MP4 files for minimum size. 

\subsection{Pedals}
The pedals of our current da Vinci model consist of \textit{camera}, \textit{clutch}, \textit{monopolar}, and \textit{bipolar} functionalities. However, the dVRK provides the pedal signals solely at the moment when the pedals are pressed. Consequently, to achieve synchronization with the image and kinematic data, we interpolated the signals for the camera and clutch pedals, where the signals are $0$ by default and remain $1$ while the pedal is pressed. This interpolation ensures coherence across the image and kinematic data.

\begin{figure}[t]
    \centering
    \includegraphics[width=\linewidth]{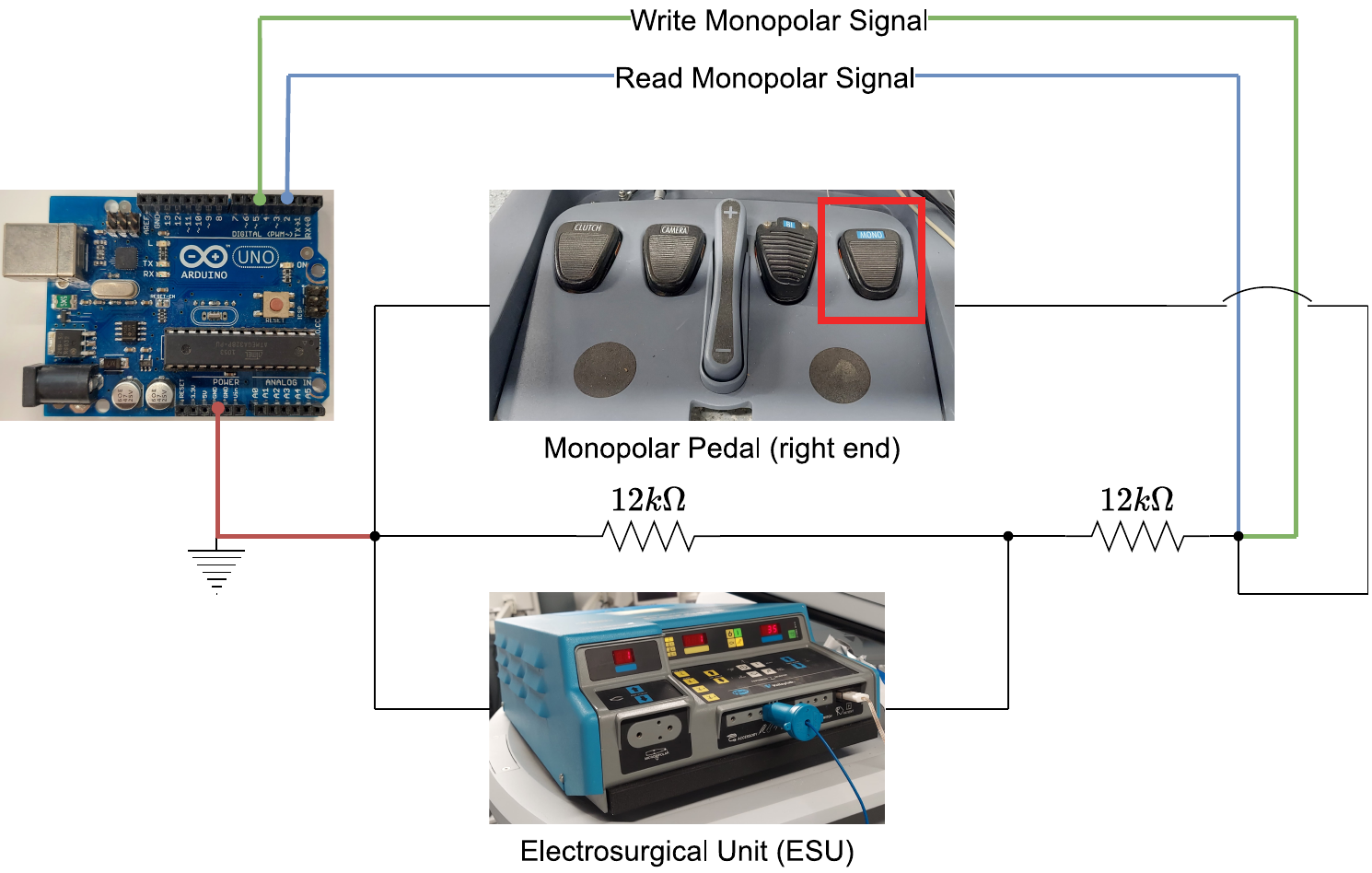}
    \caption{Circuit of the monopolar pedal and the electrosurgical generator connected to the Arduino Uno Device.}
    \label{fig:mp_circuit}
\end{figure}

\begin{figure}[t]
    \centering
    \begin{subfigure}[b]{0.8\linewidth}
        \centering
        \includegraphics[width=\linewidth]{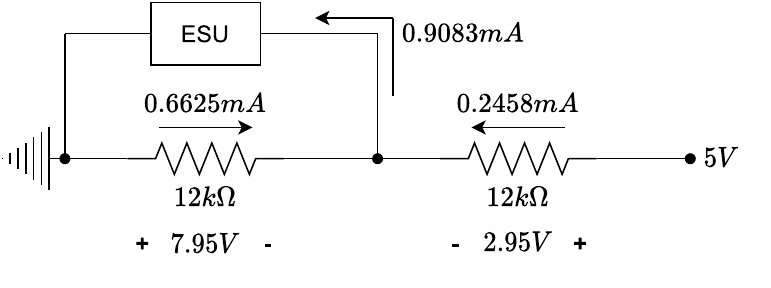}
        \caption{Circuit of Fig.~\ref{fig:mp_circuit} when it's off.}
        \label{fig:mp_off}
    \end{subfigure}
    \begin{subfigure}[b]{0.8\linewidth}
        \centering
        \includegraphics[width=\linewidth]{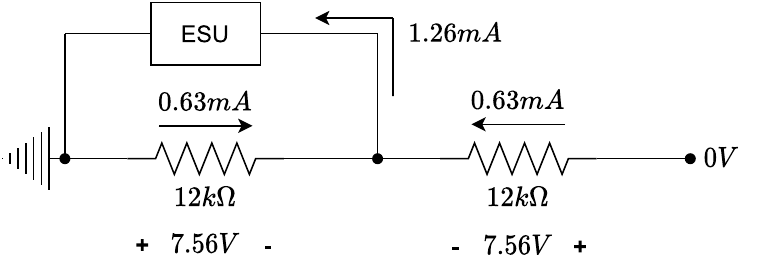}
        \caption{Circuit of Fig.~\ref{fig:mp_circuit} when it's on.}
        \label{fig:mp_on}
    \end{subfigure}
    \caption{Circuit of Fig.~\ref{fig:mp_circuit} when the monopolar output is deactivated (a), and when it is activated (b).}
    % \label{fig:mp_on_off}
\vspace{-5mm}
\end{figure}

Moreover, the dVRK lacks direct control over the electrosurgical generator (or electrosurgical unit, ESU) responsible for regulating the voltage output of the monopolar instruments used for tissue dissection. In our setup, the Pfizer Valleylab Force 2 electrosurgical generator was used, where its input schematic remained a black box. However, we discovered that a minimum current of $1 mA$ must flow through the input cable originally connected to the pedals to activate the monopolar output of the generator. To address this, we established an interface between the generator's input cable and the da Vinci console pedals using an Arduino, as depicted in Fig.~\ref{fig:mp_circuit}. The Arduino's write pin defaults to high ($5V$), and the voltage distribution is depicted in Fig.~\ref{fig:mp_off}. In this state, the current remains below the threshold, preventing activation of the monopolar output. The monopolar output is activated either when the Arduino's write pin is triggered (set to low) or when the user presses the pedal, causing the end voltage to short to $0V$, thereby surpassing the threshold as illustrated in Fig.~\ref{fig:mp_on}.

\subsection{Kinematic Dataset}

\begin{figure}[t]
    \centering
    \includegraphics[width=\linewidth]{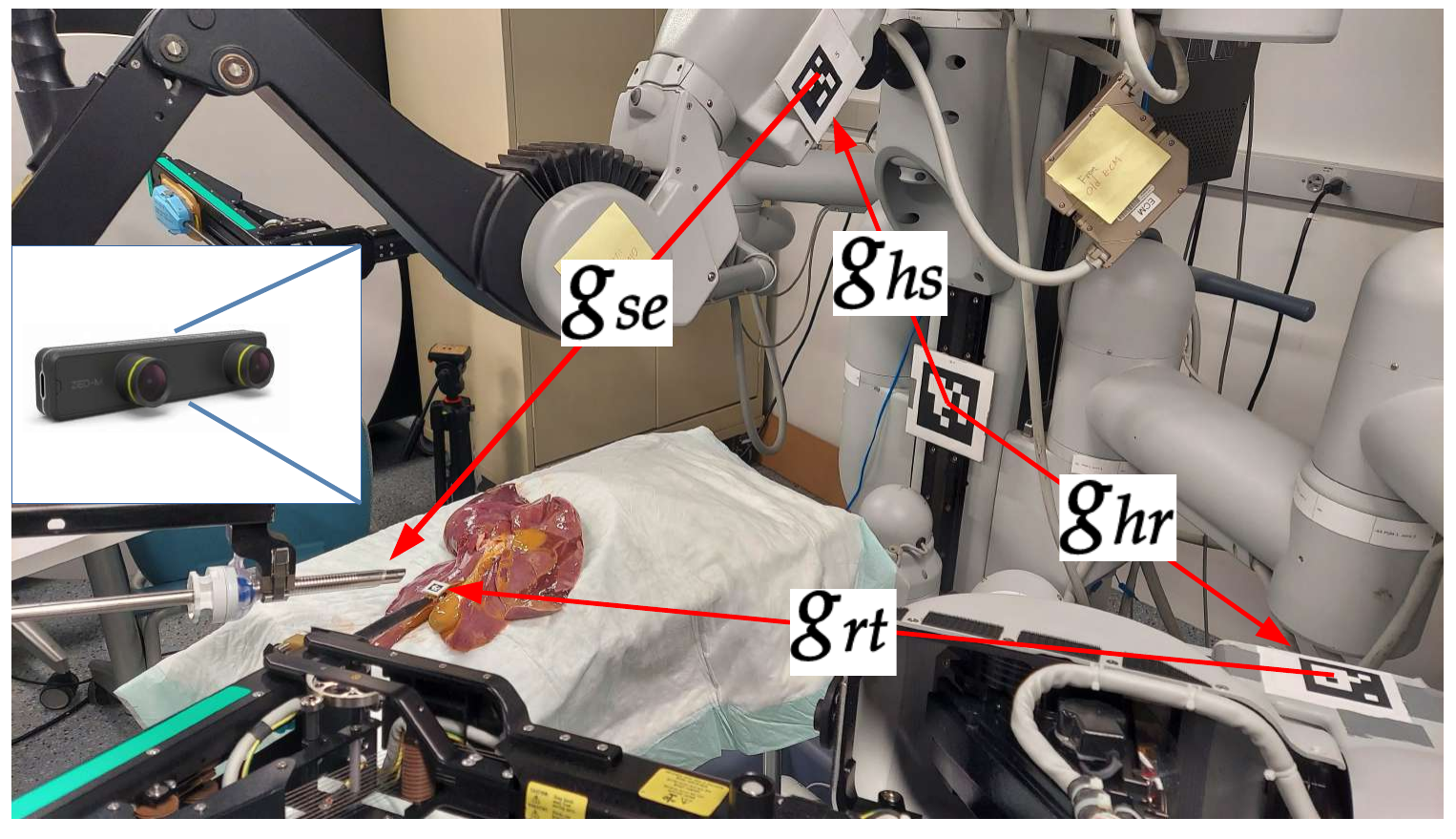}
    \caption{The setup showing how our custom-calibrated kinematics work. The transformations are shown based on the direction of the arrows and eventually, they are used to find the transformation between the ECM tip and PSM tip.}
    \label{fig:calibsetup}
\end{figure}

\begin{figure}[t]
    \begin{subfigure}[b]{0.24\textwidth}
        \centering
        \includegraphics[width=\linewidth]{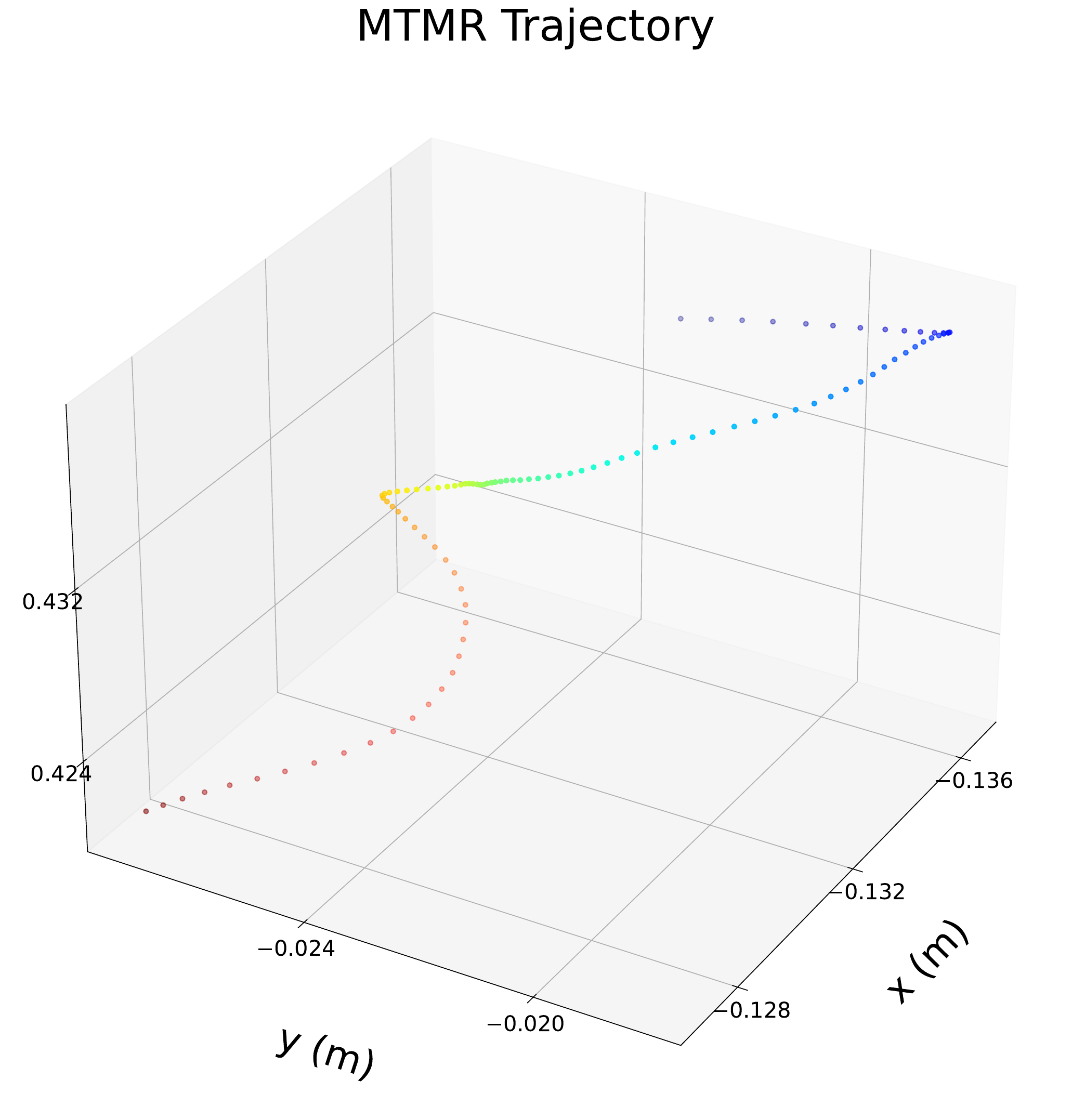}
        \caption{}
        % \label{fig:mtmr3d}
    \end{subfigure}
    \begin{subfigure}[b]{0.24\textwidth}
        \centering
        \includegraphics[width=\linewidth]{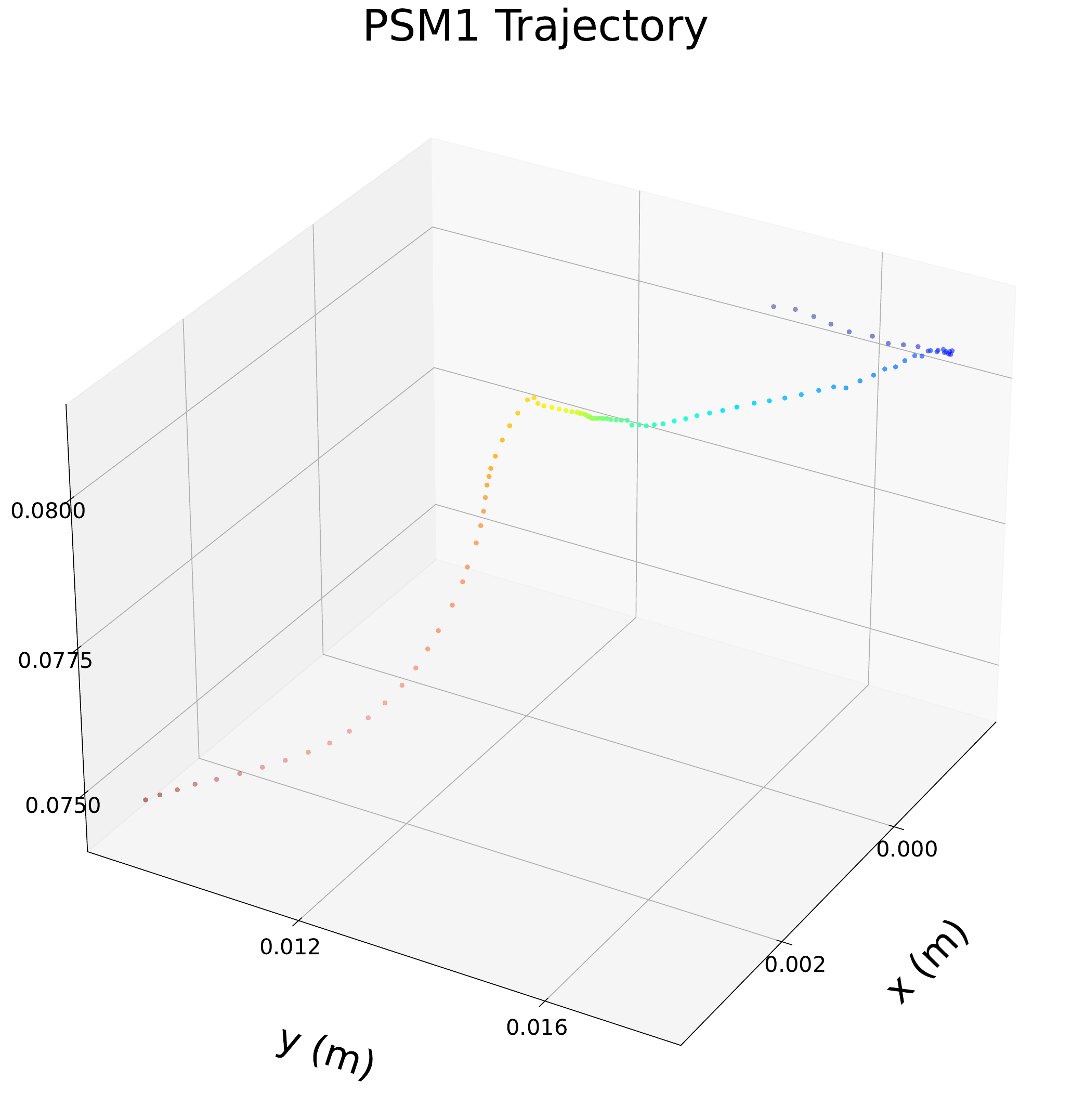}
        \caption{}
        % \label{fig:psm13d}
    \end{subfigure}
    \caption{A sample from the recorded 3D trajectory of the (a) MTMR and (b) PSM1. }
    \label{fig:3dtrajs}
\vspace{-5mm}
\end{figure}

In the full da Vinci system, the forward kinematics of the dVRK are deduced from the Setup Joints (SUJs) located at the base of the da Vinci robot. This computation enables the determination of the Patient Side Manipulator (PSM) tip's configuration relative to the Endoscope Camera Manipulator (ECM) tip. It should be noted that the positional variance between the PSM tip and the ECM tip is restricted within a range of $\pm 5cm$ for translation and falls between $5\sim10$ degrees for orientation, as reported in~\cite{dvrk}. Therefore, we introduced a custom calibration for the dVRK using fiducial markers~\cite{oh2023framework}.

This approach is influenced by~\cite{camera_calib_dvrk}, which employs an optical tracking system with custom adapters for instrument tips to calibrate the dVRK. Due to the limited availability of such a system, we opted for the more accessible ArUco markers~\cite{GARRIDOJURADO20142280}.  Furthermore, we strategically selected base frames for each arm to enhance flexibility, ensuring consistent performance during actual surgical procedures, as illustrated in Fig.~\ref{fig:calibsetup}. More information on the derivation of forward and inverse kinematics can be found in~\cite{oh2023framework}.

In Fig.~\ref{fig:calibsetup}, each $g_{ab}$ represents the transformation (homogeneous matrix) between frames $A$ and $B$. The base frames for the PSM and ECM are denoted by $R$ and $S$, respectively, while $T$ and $E$ correspond to their respective instrument tip frames. Once $g_{rt}$ and $g_{se}$ are determined, we can establish the relative configuration of the PSM tip with respect to the ECM tip, incorporating the Helper ($H$) frame as depicted in Fig.~\ref{fig:calibsetup}. If there are changes in the locations of the Setup Joints (SUJs), only the transformation between the helper and base frames requires an update. We have:
\begin{equation}
g_{et} = g_{se}^{-1}\cdot g_{hs}^{-1}\cdot g_{hr}\cdot g_{rt} = g_{es}\cdot g_{sh}\cdot g_{hr}\cdot g_{rt}
\end{equation}
For the PSMs, the dataset encompasses the following information: the transformation from the arm's base frame to its instrument tip ($g_{rt}$), the transformation from the ECM tip to the PSM instrument tip ($g_{et}$), the joint states (position, velocity, and effort) received from the dVRK, and the joint states (position, velocity, and effort) of the jaw.

Regarding the ECM, the dataset includes the transformation from the arm's base frame to its instrument tip ($g_{se}$), the transformation from the Helper frame to the ECM tip ($g_{he}$), and the joint states (position, velocity, and effort) provided by the dVRK.

For the Master Tool Manipulator (MTM), where the surgeon controls the robot, the raw kinematic data from the dVRK~\cite{kinematic_dataset, dvrk} was recorded. This data contains the transformation from the base of each arm to its controller tip, the transformation from the High-Resolution Stereo Video (HRSV) frame (where the console monitor is positioned) to the controller tip, the joint states (position, velocity, and effort) of each arm, and the joint angle of each gripper. The PSM1 is associated with the MTMR (MTM Right), while the PSM2 is paired with MTML (MTM Left). Fig.~\ref{fig:3dtrajs} illustrates a brief motion of the MTMR and the corresponding PSM1.

% \begin{table}[t]
% \centering
% \begin{tabular}{lcccc} \hline \hline\\
% \textbf{Feature}      & \textbf{Dim} & \textbf{Type}& \textbf{Description}                        \\

%                                       &                                 \\ \hline \\ 
% MTMR Transform                         &  7                     &  Float              & Right Controller Pose\\
% MTML Transform                           &  7                     &  Float              &  Left Controller Pose               \\
% ECM Transform                  &  7                     &  Float          &       Endoscope Pose          \\
% PSM1 Transform                      &  7                  &  Float              &     Arm1 Patient Side Pose            \\
% PSM2 Transform                              &  7                     &  Float              &Arm2 Patient Side Pose        \\
% Data Clutch                      &  1                    & Bool      & Clutch Pedal State       \\
% Data Camera      &  1                    & Bool      &   Camera Pedal State              \\
% Data  Monopolar         & 1                    &   Bool    & Monopolar Pedal State                \\  \\ \hline
%               \\ \hline \hline
% \end{tabular}
% \caption{Principal signal content of each feature set. Dimensions are counted for $N=3$.}
% \label{tab:feature_set_signals}
% \end{table}

\begin{table*}[t]
\centering
\renewcommand{\arraystretch}{1.5}
\begin{tabular}{l|c|c|c}
\hline
\hline
\textbf{Type} & \textbf{Features} & \textbf{Dim} & \textbf{Description (Units)} \\ \hline
\multirow{3}{*}{\textbf{ECM}}       & Endoscope Tip Cartesian Pose         & 7       & Translation \{$x$, $y$, $z$\} (m), Quaternion \{$x$, $y$, $z$, $w$\} (rad) \\ \cline{2-4}
                                    & Local Endoscope Tip Cartesian Pose   & 7       & Translation \{$x$, $y$, $z$\} (m), Quaternion \{$x$, $y$, $z$, $w$\} (rad) \\ \cline{2-4}
                                    & Arm Joint State                      & 12      & Joint Position \{$\theta(4)$\} (rad), Joint Velocity \{$\dot{\theta}(4)$\} (rad/s), Joint Effort \{$\tau(4)$\} (N) \\ \hline
                                    
\multirow{4}{*}{\textbf{MTML}}      & Manipulator Tip Cartesian Pose       & 7       & Translation \{$x$, $y$, $z$\} (m), Quaternion \{$x$, $y$, $z$, $w$\} (rad) \\ \cline{2-4}
                                    & Local Manipulator Tip Cartesian Pose & 7       & Translation \{$x$, $y$, $z$\} (m), Quaternion \{$x$, $y$, $z$, $w$\} (rad) \\ \cline{2-4}
                                    & Manipulator Joint State              & 18      & Joint Position \{$\theta(6)$\} (rad), Joint Velocity \{$\dot{\theta}(6)$\} (rad/s), Joint Effort \{$\tau(6)$\} (N) \\ \cline{2-4}
                                    & Manipulator Gripper Joint State      & 1       & Joint Position \{$\theta(1)$\} (rad) \\ \hline
                                    
\multirow{4}{*}{\textbf{MTMR}}      & Manipulator Tip Cartesian Pose       & 7       & Translation \{$x$, $y$, $z$\} (m), Quaternion \{$x$, $y$, $z$, $w$\} (rad) \\ \cline{2-4}
                                    & Local Manipulator Tip Cartesian Pose & 7       & Translation \{$x$, $y$, $z$\} (m), Quaternion \{$x$, $y$, $z$, $w$\} (rad) \\ \cline{2-4}
                                    & Manipulator Joint State              & 18      & Joint Position \{$\theta(6)$\} (rad), Joint Velocity \{$\dot{\theta}(6)$\} (rad/s), Joint Effort \{$\tau(6)$\} (N) \\ \cline{2-4}
                                    & Manipulator Gripper Joint State      & 1       & Joint Position \{$\theta(1)$\} (rad) \\ \hline
                                    
\multirow{4}{*}{\textbf{PSM1}}      & Instrument Tip Cartesian Pose        & 7       & Translation \{$x$, $y$, $z$\} (m), Quaternion \{$x$, $y$, $z$, $w$\} (rad) \\ \cline{2-4}
                                    & Local Instrument Tip Cartesian Pose  & 7       & Translation \{$x$, $y$, $z$\} (m), Quaternion \{$x$, $y$, $z$, $w$\} (rad) \\ \cline{2-4}
                                    & Arm Joint State                      & 18      & Joint Position \{$\theta(6)$\} (rad), Joint Velocity \{$\dot{\theta}(6)$\} (rad/s), Joint Effort \{$\tau(6)$\} (N) \\ \cline{2-4}
                                    & Instrument Tip Jaw Joint State       & 3       & Joint Position \{$\theta(1)$\} (rad), Joint Velocity \{$\dot{\theta}(1)$\} (rad/s), Joint Effort \{$\tau(1)$\} (N) \\ \hline
                                    
\multirow{4}{*}{\textbf{PSM2}}      & Instrument Tip Cartesian Pose        & 7       & Translation \{$x$, $y$, $z$\} (m), Quaternion \{$x$, $y$, $z$, $w$\} (rad) \\ \cline{2-4}
                                    & Local Instrument Tip Cartesian Pose  & 7       & Translation \{$x$, $y$, $z$\} (m), Quaternion \{$x$, $y$, $z$, $w$\} (rad) \\ \cline{2-4}
                                    & Arm Joint State                      & 18      & Joint Position \{$\theta(6)$\} (rad), Joint Velocity \{$\dot{\theta}(6)$\} (rad/s), Joint Effort \{$\tau(6)$\} (N) \\ \cline{2-4}
                                    & Instrument Tip Jaw Joint State       & 3       & Joint Position \{$\theta(1)$\} (rad), Joint Velocity \{$\dot{\theta}(1)$\} (rad/s), Joint Effort \{$\tau(1)$\} (N) \\ \hline \hline
\multirow{3}{*}{\textbf{Pedals}}    & Clutch Pedal State                  & 1       & Boolean (True when activated, False otherwise)  \\ \cline{2-4}
                                    & Camera Pedal State                  & 1       & Boolean (True when activated, False otherwise) \\ \cline{2-4}
                                    & Monopolar Pedal State               & 1       & Boolean (False when activated, True otherwise) \\
\hline
\hline
\end{tabular}
\caption{List of kinematic variables, including the pedal signals. ``Local'' tip cartesian pose relates the position of the arm's tip to the arm's base frame (e.g., $g_{rt}$ of the PSM). Otherwise, it is the pose of the arm's tip to its reference frame (the Helper frame, ECM tip frame, and HRSV frame for ECM, PSMs, and MTMs, respectively).}
% \label{tab:feature_set_signals}
\vspace{-5mm}
\end{table*}

\section{Surgical Task} \label{section:task}

\subsection{Setup and repetitions}

\begin{table}[t]
\centering
\begin{tabular}{c|c|c|c}
\hline
\hline
               & \textbf{Video} & \textbf{Kinematics} & \textbf{Pedals} \\ \hline
\textbf{A}     & 1              & 3                   & 3               \\
\textbf{B}     & 3              & 3                   & 3               \\
\textbf{C}     & 3              & 3                   & 3               \\
\textbf{D}     & 0              & 3                   & 3               \\
\textbf{E}     & 3              & 3                   & 3               \\
\textbf{F}     & 3              & 3                   & 0               \\
\textbf{G}     & 3              & 3                   & 3               \\  \hline

\textbf{Total} & 16             & 21                  & 18              \\  
\hline
\hline
\end{tabular}
\caption{Recorded dataset distribution from each subject. Some recordings are excluded due to corruption.}
\label{tab:surgdata}
\vspace{-5mm}
\end{table}

\begin{figure}[t]
    \centering
    \includegraphics[width=\linewidth]{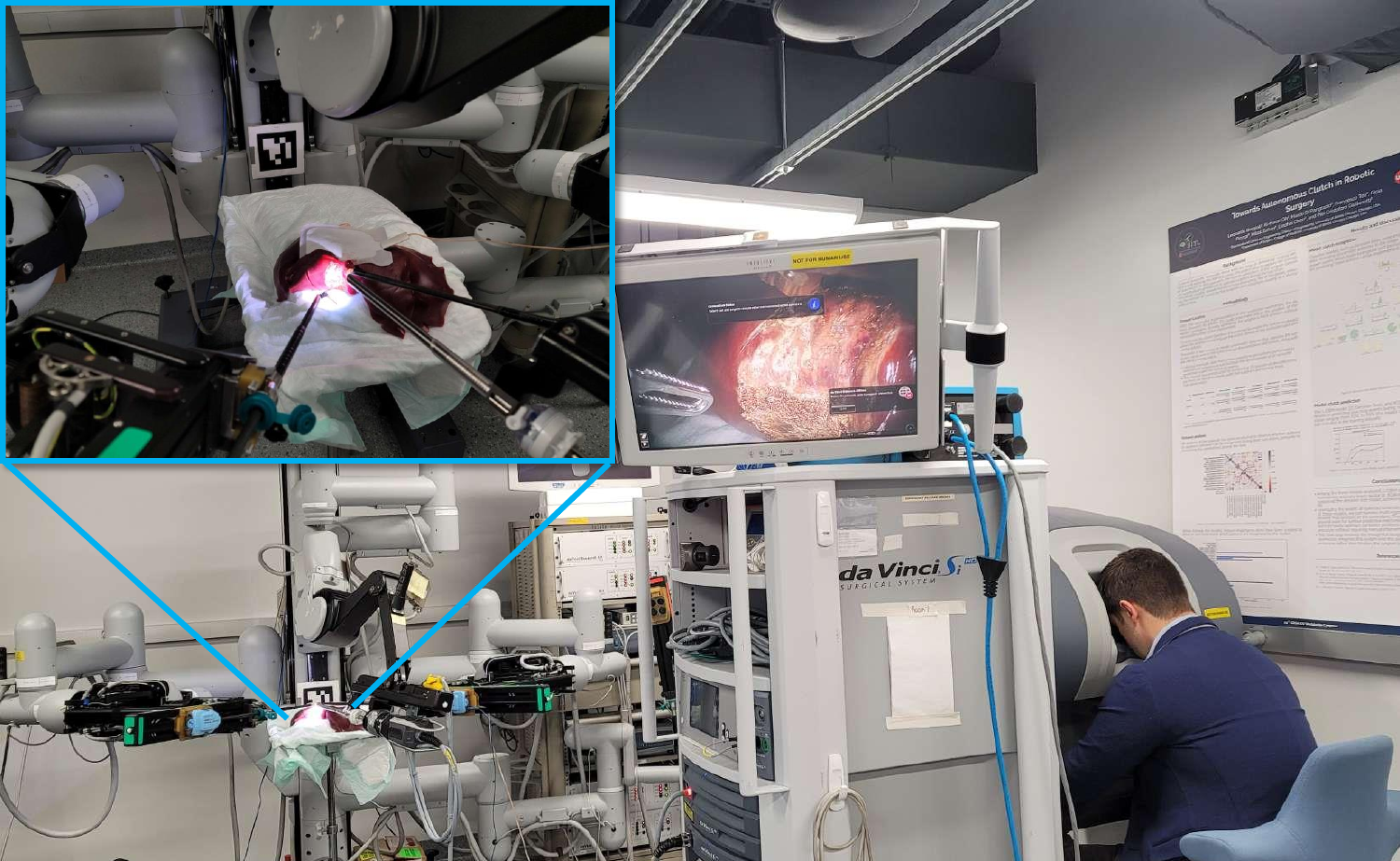}
    \caption{The environment setup for the ex-vivo cholecystectomy performed by a surgeon.}
    \label{fig:envsetup}
\vspace{-5mm}
\end{figure}

The recordings took place in the setup depicted in Fig.~\ref{fig:envsetup}, where the surgeon controls the robot with the da Vinci console and executes the assigned task on a pig liver. The data set comprises seven surgeons denoted alphabetically from ``A'' to ``G'' who all have experience in surgical robotic cholecystectomy. Each subject performed the task three times. The duration of the task varied according to the difficulty level of the task, influenced by factors such as the decay of the liver. In particular, challenges arose when the color similarity between the liver and gallbladder increased, making it difficult to distinguish between the two objects. 

Table~\ref{tab:surgdata} provides details on the recorded dataset for each surgeon. Note that some videos were damaged during compression and were consequently excluded from the dataset. Additionally, occasional shutdowns of the Arduino occurred when a high current was applied to the instrument tip, resulting in the corruption of pedal recordings.

\subsection{Description} \label{section:task_desc}
The surgeons performed the task following the UIC standardized surgical technique for robotic cholecystectomy~\cite{giulianotti2024foundation}. It is worth noting that the order of certain steps may potentially vary based on the specific surgical case or anatomical considerations. The primary steps of the procedure are as follows:

\begin{enumerate}
    \item Working area exposure
    \item Gallbladder neck retraction
    \item Calot triangle: anterior peritoneal layer opening
    \item Calot triangle: posterior peritoneal layer opening
    \item Cystic duct isolation
    \item Cystic artery isolation
    \item Cystic duct clipping
    \item Cystic artery clipping
    \item Cystic duct and artery division
    \item Detachment of the gallbladder from the liver
    \item Specimen retrieval in an Endobag\textsuperscript{TM}
\end{enumerate}

However, certain simplifications were applied to the technique mentioned above for this study and within the context of this experimental animal model. In particular, steps 8, 9, and 11 were omitted.

\section{Preliminary Work}

\subsection{Pedal Prediction}

%The first task was ...
In the context of robotic cholecystectomy, predicting the surgeon's actions, particularly those involving clutching and manipulating camera pedals, is essential to optimize procedural efficiency and alleviate the surgeon's cognitive workload. In the current setting, the surgeon takes full control of the robotic system without direct collaboration. Nevertheless, there exists potential to develop a classifier leveraging a comprehensive dataset encompassing the robot's kinematics and pedal signals. This classifier holds the capability to predict the surgeon's actions and, subsequently, can be integrated into control systems to guide how the robot should respond.

\subsubsection{Dataset Preprocessing}

As indicated in Table~\ref{tab:surgdata}, the robot kinematics ($\sim100$Hz) and the console pedal inputs ($\sim230$Hz) were captured independently due to their varying frequencies. This necessitated a synchronization process before training classifiers. The initial step involved aligning the datasets by comparing their timestamps and match entries that exhibited a discrepancy within a threshold of $\epsilon=0.006s$. Subsequently, to ensure a uniform data quantity, the kinematics data, which was recorded at a lower frequency, underwent interpolation to match the volume of pedal data.

\renewcommand{\arraystretch}{1.4}
\begin{table}[t]
\centering
\begin{tabular}{l|c|c|c|c}
\hline\hline
\textbf{Pedal Type}              & \textbf{Data Type} & \textbf{Not Pressed} & \textbf{Pressed} & \textbf{Ratio} \\ \hline \hline
\multirow{2}{*}{\textbf{Clutch}} & Original           & 205406               & 934              & 220            \\ \cline{2-5} 
                                 & Undersampled       & 18700                & 934              & 20             \\ \hline
\multirow{2}{*}{\textbf{Camera}} & Original           & 201125               & 5216             & 36             \\ \cline{2-5} 
                                 & Undersampled       & 78240                & 5216             & 20             \\ \hline \hline
\end{tabular}
\caption{Class distribution before and after undersampling}
\label{tab:class_distribution}
\end{table}

\renewcommand{\arraystretch}{1.1} % Adjust the factor as needed
\begin{table}
\centering
\begin{tabular}{l|l|c|c}
\hline \hline 
\textbf{Models}   & \textbf{Evaluation metrics}  & \textbf{Clutch} & \textbf{Camera} \\  \hline \hline
                                   
\multirow{3}{*}{\textbf{AdaBoost}} & Precision                           & 0.8252  & 0.9104 \\
                                   & F1 score                              & 0.7540 & 0.9145 \\
                                   & Recall                           & 0.6941 & 0.9187 \\ \hline 
                                   
\multirow{3}{*}{\textbf{NN}}       & Precision                           & 0.8209 & 0.9230 \\
                                   & F1 score                              & 0.6688 & 0.7467 \\
                                   & Recall                            & 0.5354 & 0.6270  \\ \hline 
                          
\multirow{3}{*}{\textbf{Random Forest}}       & Precision                           & 0.8726 & 0.9117 \\
                                              & F1 score                              & 0.8379 & 0.9177 \\
                                              & Recall                           & 0.8059 & 0.9137 \\  \hline 
                          
\multirow{3}{*}{\textbf{L-GBM}}       & Precision                            & 0.8503 & 0.9147 \\
                                      & F1 Score                              & 0.8427 & 0.9256 \\
                                      & Recall                            & 0.8353 & 0.9367 \\  \hline \hline
\end{tabular}
\caption{Accuracy, Recall, and F1 scores of each model measured on the test set.}
\label{tab:evalmetrics}
\vspace{-5mm}
\end{table}

Upon activating the clutch or the camera pedal, the orientations of both the robot arms (PSMs or ECM) and the manipulators (MTMs) are locked in place. However, during this state, the manipulators retain the ability to move while the positions of the da Vinci arms remain stationary. Consequently, we used a sliding window of 20 sequential instances ($\sim0.6s$) and measured the travel distance associated with the Cartesian translations of the arms and the manipulators. These calculated distances served as the input features for the classifiers, reflecting their distinct behaviors and constraints in response to pedal activation.

The labels for the selected features, which correspond to the pedal signal, exhibit imbalance as highlighted in Table~\ref{tab:class_distribution}. This imbalance stems from the pedals being idle (unpressed) predominantly during the procedure. To mitigate the model's potential bias towards the majority class, we experimented with various ratios for balancing the classes, \textit{True} when the pedal is pressed and \textit{False} otherwise, ranging from 2 to 30, with increments of 2. Through this process, we determined that a ratio of 20 produced the most favorable outcomes in model performance metrics. 

% Finally, the dataset is randomly split into training and test sets with an $8:2$ ratio.

% Addressing this imbalance in binary classification is crucial to prevent model bias towards the majority class, ensuring accurate evaluation, and preventing the loss of critical information, especially when the minority class is of interest. This would result in models never detecting the positive class. To tackle these challenges, we chose to randomly undersample the dataset when the pedal is not in use (\textit{False}), maintaining a ratio of 20 compared to when the pedal is in use (\textit{True}). It is important to note that the dataset still needs to represent the rarity of a clutch event accurately compared to the overall procedure. For instance, a ratio of 2 would lead to overly sensitive models that generate too many false positives. Ratios ranging from 2 to 30 were tested (with an increment of 2), and the one yielding the best scores for the models was selected.

\subsubsection{Trained Classifiers}

The Random Forests (RF)~\cite{breiman2001random}, Adaptive Boosting (AdaBoost)~\cite{freund1997decision}, and Feedforward Neural Networks (FNN)~\cite{Goodfellow-et-al-2016} classifiers employed in this study are analogous to those introduced in~\cite{rysbek2023intent}. However, the models have been specifically tuned for binary classification. Furthermore, the structure of the FNN consists of an input layer ($n_{0}$) sized according to the input characteristics, followed by two fully connected layers ($n_{1}$, $n_{2}$), where the number of neurons is proportional to the size of the previous layer ($n_{1} = 2\cdot n_{0}$, $n_{2} = 2/3\cdot n_{1}$). The final layer ($n_{f}$) has $1$ output to function as a binary classifier.

Furthermore, we incorporated a novel classifier, the Light Gradient Boosting Machine (L-GBM)~\cite{NIPS2017_6449f44a}. In L-GBM, decision trees are sequentially added, each tree correcting errors from the ensemble of preceding trees. Unlike conventional gradient boosting, L-GBM directly minimizes the loss function. Designed for distributed and efficient training on large datasets, it is well-suited for applications with resource constraints. L-GBM is distinguished for its leaf-wise growth strategy, effective handling of categorical features, and parallel/distributed training capabilities.

\subsubsection{Model Evaluations}

The entirety of the preprocessed window-level datasets, as outlined in Table~\ref{tab:class_distribution}, has been divided into training and test sets, allocated at proportions of $80\%$ and $20\%$, respectively. The overall performance of each trained model on the window is described in Table~\ref{tab:evalmetrics}, including the precision, recall, and F1 score. 

\begin{figure}[t]
    \centering
    % \includesvg[width=\linewidth]{figures/pred_plot.svg}
    \includegraphics[width=\linewidth]{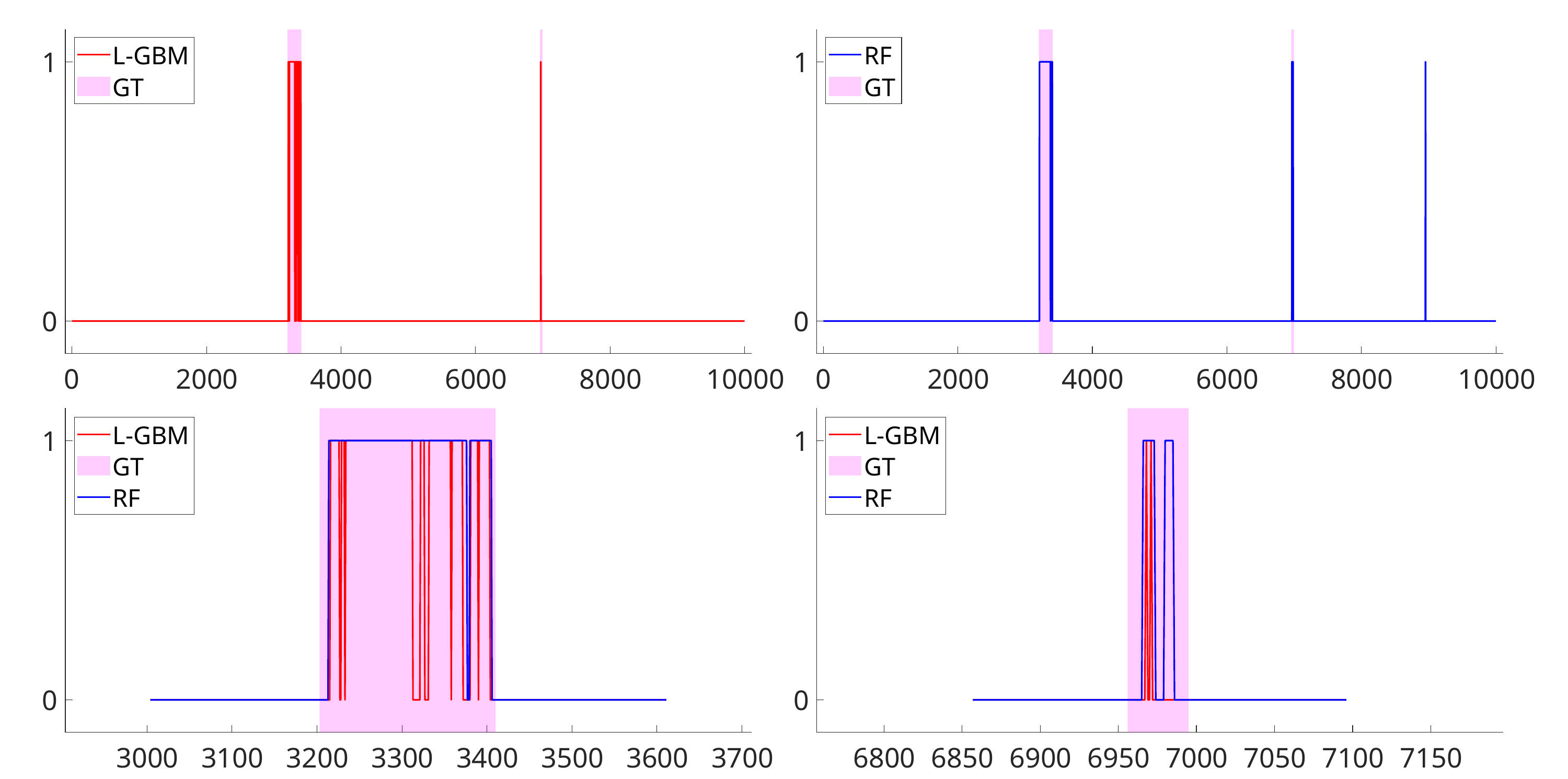}
    \caption {Models evaluation on the total dataset. The top figures show the predictions from two models (L-GBM and Random Forest). The bottom figures present a zoomed view of the top figure where the clutch was activated.}
    \label{fig:predictions_graph}
    \vspace{-5mm}
\end{figure}

We considered accuracy at the signal level to further assess model performance beyond window-level metrics. Accordingly, we excluded one of the pedal-kinematic procedures outlined in Table~\ref{tab:surgdata} from the dataset. From Table~\ref{tab:evalmetrics}, the RF and the L-GBM outperformed FNN and AdaBoost for predicting clutch. Thus, we selected these models for signal-level performance evaluation. In Fig.~\ref{fig:predictions_graph}, we exhibited a trade-off between the model's sensitivity and its propensity to generate false positives. RF was more stable across long clutch periods but predicted a false clutch at the end of the dataset.
In contrast, L-GBM showed precise clutch activation predictions but was unstable over prolonged clutch activations. Further processing could resolve this limitation, such as the voting technique~\cite{varol_multiclas_realtime_intent}. This method introduces fixed time delays depending on the voting buffer size; however, it enhances the accuracy and reduces short-term false transitions.

% Finally, by training models like L-GBM, it became feasible to predict the engagement of the clutch pedal. As demonstrated in Table~\ref{tab:evalmetrics},  We conducted tests on a procedure not included in the training set to validate their precision. However, only the L-GBM classifier could deliver highly accurate predictions when the model underwent this specific testing, as illustrated in Fig.~\ref{fig:predictions_graph}. There is a trade-off between the model's sensitivity and the tendency to create false positives for these classifiers. For example, Random Forest is more stable during long instances of a clutch but predicts a clutch at the end of the dataset. Further processing could correct this by creating, for example, a voting window. Finally, the goal of this implementation was to prove the efficiency of the dataset in creating models, the goal that we consider achieved.

\subsection{Training Tissue Segmentation Models}

% Another example of how the new dataset has been used is ...
To automate partial aspects of robotic cholecystectomy, the robot must recognize and keep track of the tissues during the procedure. Currently, there is a notable scarcity of datasets specifically designed for such research. However, this new dataset, in contrast to existing ones, uniquely captures the dynamic changes in tissues during cholecystectomy procedures. Notably, the tissues exhibit a rich diversity in both shape and color. This deliberate inclusion of diverse tissue characteristics is pivotal for training segmentation models, enabling the robot to adeptly recognize and track tissues in real-time during surgical procedures.

\subsubsection{Generating segmentation dataset}

\begin{figure}[t]
    \centering
    \includegraphics[width=\linewidth]{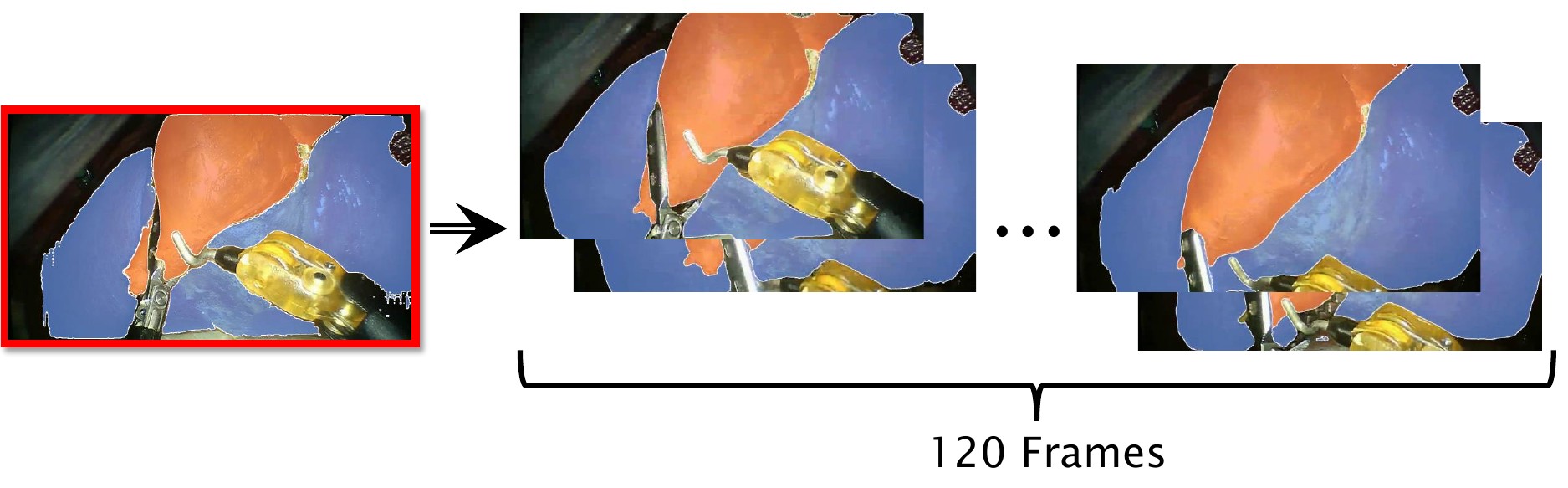}
    \caption{An example of generating annotations with Track-Anything. Once the initial frame of the video clip (red box) is annotated, Track-Anything starts annotating the rest of the frames.}
    \label{fig:trackany}
\vspace{-5mm}
\end{figure}

In~\cite{oh2023framework}, a custom dataset featuring a pig's liver and gallbladder was generated, and it was used to train an object segmentation model called Detectron2~\cite{wu2019detectron2}. However, this dataset shares similarities with~\cite{colleoni2020synthetic}, wherein the arms and the endoscope were manually moved around the object, but no actions were performed on the tissues. Moreover, this elementary dataset was recorded on a single tissue. This posed a significant limitation when it was tested on a new tissue or as the shapes and colors of the tissues changed when the energy was delivered. Consequently, the previously trained model encountered challenges in real-time tissue recognition during the automated procedure. Moreover, the dataset's size is notably low compared to modern datasets, as video frames had to be downsampled, and each frame had to be manually annotated.

We addressed the limitations inherent in the existing dataset by generating a new dataset annotated using Track Anything (TA)~\cite{yang2023track}. Surgical videos from two distinct surgeons (E and F) were utilized, where one depicted a nearly ideal cholecystectomy, and the other showed a procedure in a challenging surgical environment. This selection aimed to expose the model to diverse surgical scenarios, ensuring it learns to accurately handle complex situations in actual surgeries. One limitation of TA is its computational efficiency, which decreases as the number of frames in a video increases. Consequently, we split the videos into short clips with a duration of $2$ seconds (equivalent to $120$ frames). After annotating the first frame, TA automatically extends the annotations to the remaining frames (Fig.~\ref{fig:trackany}). Adjustments to the auto-annotated results could be made if necessary. The training set included a total of $50,149$ annotated images (approximately 35 times from~\cite{oh2023framework}).
% , leaving $12,538$ images for evaluation 

\subsubsection{Training Results}

Table~\ref{tab:dt2res} compares the Average Precision (AP)~\cite{mscoco} results for two models, each trained separately on our initial dataset (from~\cite{oh2023framework}) and the recently generated dataset. The models were evaluated on an independent dataset of 1104 images from one of Participant C's videos. The previous model trained on a controlled dataset shows markedly limited performance on surgical images, indicated by its AP scores. In contrast, the new model, even when trained on just a fraction (two out of 16 videos) of actual surgical footage, demonstrates an improvement over the previous model's performance. Despite the overall modest scores, the new model's superiority in this practical context reflects its enhanced capacity to adapt to the complexities of genuine cholecystectomy procedures, hinting at a significant potential for further improvements with comprehensive training on the remaining surgical videos.

\renewcommand{\arraystretch}{1.1}
\begin{table}[t]
\centering
\begin{tabular}{l|l|c|c}
\hline
\hline
                                         & \textbf{Categories} & \textbf{AP (Box)} & \textbf{AP (Seg.)} \\ \hline
\multirow{2}{*}{\textbf{Previous Model}} & Pig Liver           & 33.3              & 16.0               \\ \cline{2-4} 
                                         & Pig Gallbladder     & 11.4              & 10.7               \\ \hline
\multirow{2}{*}{\textbf{New Model}}      & Pig Liver           & 62.3              & 51.9               \\ \cline{2-4} 
                                         & Pig Gallbladder     & 53.9              & 49.4               \\ \hline \hline
\end{tabular}
\caption{The Average Precision (AP) scores (percentages) for each category (Box stands for Bounding Box and Seg. for Segmentation).}
\label{tab:dt2res}
\vspace{-5mm}
\end{table}

\section{Conclusion}

%The \textbf{[name of the dataset]} dataset is a surgical robotics collection of cholecystectomy recorded with the da Vinci Research Kit. The data presents (with some specified exceptions): full kinematic data on the patient side, all the pedal inputs used during the procedure, and the timestaped record of the endoscope. This new type of dataset can allow the creation of new models, two of them have been presented: one allowing the prediction of clutch and the other one predicting the use of the camera.

The current state-of-the-art primarily relies on video-annotated data from well-known datasets. However, a notable gap exists in the incorporation of kinematic data within these datasets, a limitation addressed by recent contributions such as the dataset highlighted in \cite{kinematic_dataset}. Despite these advances, challenges persist, including incomplete recordings, imprecise kinematic data due to calibration issues, and the reliance on exercise-based scenarios rather than actual procedures.

Our newly introduced dataset, recorded during ex vivo pseudo-cholecystectomy procedures on pig livers with contributions from seven surgeons, stands out by encompassing patient-side kinematic data, pedal states, and time-stamped videos. Looking ahead, future perspectives in this evolving field involve the development of advanced models to automate various subtasks during surgery, leveraging the unique attributes of data sets like ours.

To demonstrate the practical application of our dataset, we focused on predicting specific subtasks within robotic surgery scenarios. These tasks include predicting clutch and camera pedal activations in conjunction with the kinematics of the robotic arm, console manipulators, and the state of the console pedals. Within this framework, we developed two predictive models: one for clutch usage and another for camera activation, showcasing the dataset's potential to improve automation and analysis of surgical procedures.

% As another example of the utility of the dataset, we studied segmentation ...
As another example of the utility of the dataset, we studied the segmentation models' performance, crucial for the robot's ability to recognize and track tissues during cholecystectomy. The results underscore the significance of our dataset in enhancing the robot's tissue recognition capabilities. The dynamic changes captured in the tissues during cholecystectomy procedures contribute to improved training models for real-time tissue recognition. This understanding of tissue dynamics lays a foundation for advancing automation in robotic cholecystectomy, where precise tissue identification is paramount.

\clearpage
\nocite{*}
\bibliographystyle{ieeetr}
\bibliography{ismr_2024}

\end{document}